\documentclass[letterpaper, 10 pt, conference]{ieeeconf}  

\IEEEoverridecommandlockouts                              

\title{
\vspace{-60pt}
\tikzstyle{background rectangle}=[thin,draw=black]
\begin{tikzpicture}[show background rectangle]

\node[align=center, text width=0.9\textwidth, font =\small]{
This paper has been accepted for publication in \textit{IEEE ICRA 2025.} Please cite as: \\ L. Ervin, H. Bezawada, and V. Vikas, "Improving Grip Stability Using Passive Compliant Microspine Arrays for Soft Robots in Unstructured Terrain," in \textit{IEEE International Conference on Robotics and Automation (ICRA)}, 2025.
};
\end{tikzpicture} 

\LARGE \bf
Improving Grip Stability Using Passive Compliant Microspine Arrays for Soft Robots in Unstructured Terrain}

\author{Lauren Ervin, Harish Bezawada, and Vishesh Vikas$^{1}$
\thanks{*This work was supported in part by USDA/NIFA Award {\#2023-67022-40918}. The material contained in this document is based upon work supported in part by a National Aeronautics and Space Administration (NASA) grant or cooperative agreement. Any opinions, findings, conclusions, or recommendations expressed in this material are those of the authors and do not necessarily reflect the views of NASA. This work was supported through a NASA grant awarded to the Alabama/NASA Space Grant Consortium.}
\thanks{$^{1}$Lauren Ervin, Harish Bezawada, and Vishesh Vikas are with the Agile Robotics Lab, University of Alabama, Tuscaloosa, AL 35487, USA
        {\tt\small \{lefaris, hbezawada\}@crimson.ua.edu, vvikas@ua.edu}}%
}

\usepackage{graphics} 
\usepackage{epsfig} 
\usepackage{mathptmx} 
\usepackage{times} 
\usepackage{amsmath} 
\usepackage{amssymb}  
\usepackage{bm,soul,xcolor}
\usepackage{cite}
\usepackage{tikz}
\usetikzlibrary{arrows.meta, automata, positioning, quotes, backgrounds}
\usepackage{comment}
\usepackage[linesnumbered,ruled,vlined]{algorithm2e}
\usepackage{graphicx}
\usepackage{subcaption}
\usepackage{caption}
\usepackage{mathtools}
\usepackage{gensymb}
\usepackage[hidelinks]{hyperref}
\usepackage{stfloats}
\graphicspath{{Figures/}}
\newcommand{\Fig}{Fig. }

\usepackage{amssymb}

\usepackage[export]{adjustbox}

\begin{document}

\maketitle

\begin{abstract}
Microspine grippers are small spines commonly found on insect legs that reinforce surface interaction by engaging with asperities to increase shear force and traction. 
An array of such microspines, when integrated into the limbs or undercarriage of a robot, can provide the ability to maneuver uneven terrains, traverse inclines, and even climb walls. 
Meanwhile, the conformability and adaptability of soft robots makes them ideal candidates for applications involving traversal of complex, unstructured terrains. 
However, there remains a real-life realization gap for soft locomotors pertaining to their transition from controlled lab environment to the field that can be bridged by improving grip stability through effective integration of microspines.  
In this research, a passive, compliant microspine stacked array design is proposed to enhance the locomotion capabilities of mobile soft robots. A microspine array integration method effectively addresses the stiffness mismatch between soft, compliant, and rigid components. Additionally, a reduction in complexity results from actuation of the surface-conformable soft limb using a single actuator. The two-row, stacked microspine array configuration offers improved gripping capabilities on steep and irregular surfaces. This design is incorporated into three different robot configurations - the baseline without microspines and two others with different combinations of microspine arrays.
Field experiments are conducted on surfaces of varying surface roughness and non-uniformity - concrete, brick, compact sand, and tree roots.  
Experimental results demonstrate that the inclusion of microspine arrays increases planar displacement an average of 10 times.  The improved grip stability, repeatability, and, terrain traversability is reflected by a decrease in the relative standard deviation of the locomotion gaits.

\end{abstract}
\section{INTRODUCTION} 
\begin{figure}[h]
    \centering
    \includegraphics[width=\linewidth]{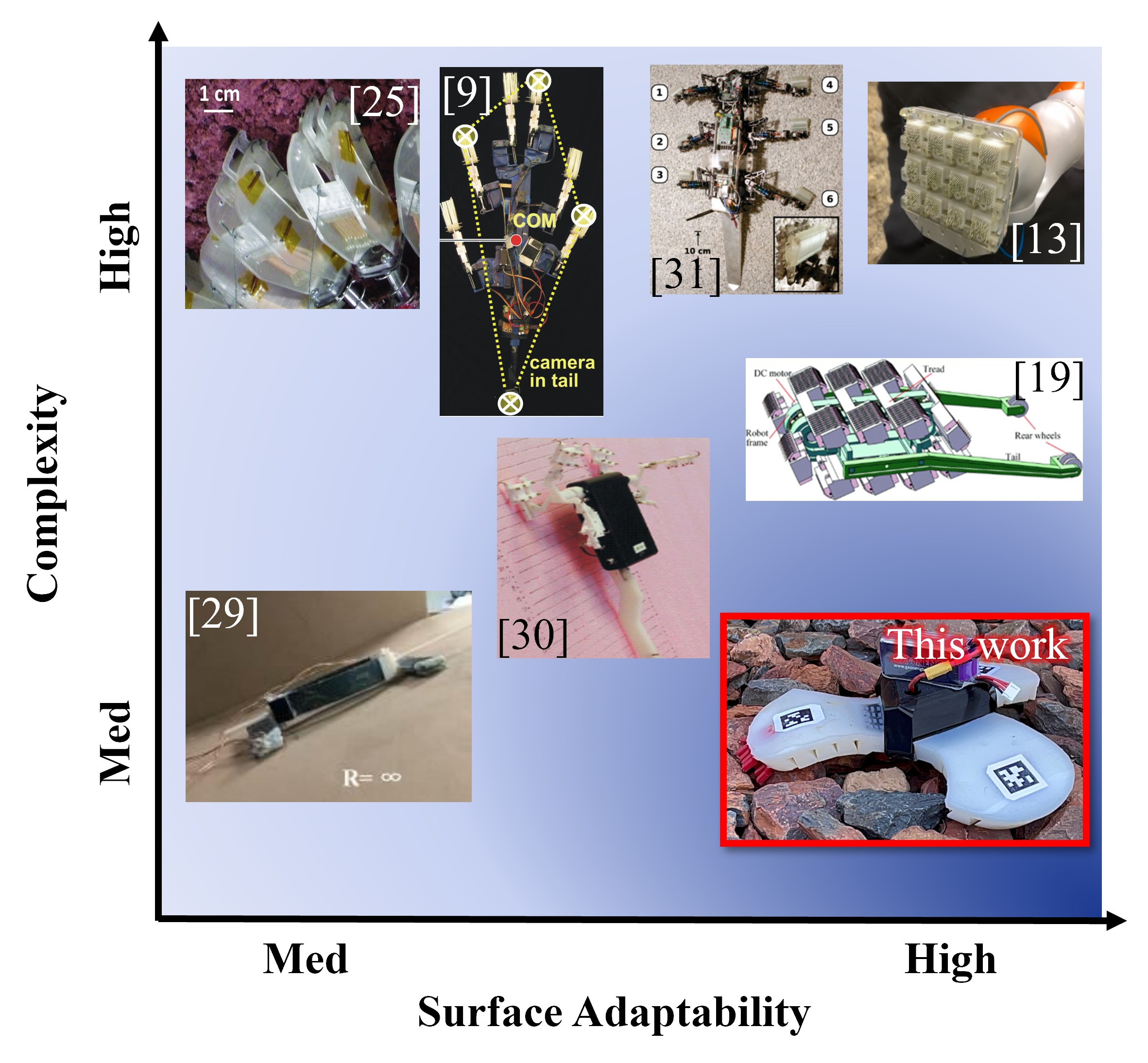}
    \caption{Surface adaptability vs. design complexity with various microspine robots \cite{asbeck_scaling_2006, wang_palm_2016, parness_lemur_2017, liu_novel_2019, hu_inchworm-inspired_2019,Daltorio_Wei_Horchler_Southard_Wile_Quinn_Gorb_Ritzmann_2009, Spenko_Haynes_Saunders_Cutkosky_Rizzi_Full_Koditschek_2008} compared against one of the presented designs in this work, 1ML.}
    \label{fig:adapt}
\end{figure}

In nature, animals resist slipping and falling by increasing interaction with surfaces in a number of ways. Snakes maintain a large surface area in contact with the ground to increase the amount of propulsive force aiding in locomotion. Snake inspired robots are developed with textured skins for increased traction \cite{liu_kirigami_2019, mckenna_toroidal_2008}. However, the efficacy is unknown for limbed robots equipped with such skins given the directional nature of the robot-environment interaction resulting from the designs. In constrast, geckos utilize van der Waals interactions to provide an adhesive force against surfaces. Consequently, attaching gecko inspired, synthetic adhesives to extremities of robots has been investigated on range of surfaces \cite{chen_testing_2022, hajj-ahmad_grasp_2023, han_climbing_2022, sangbae_kim_smooth_2008, sikdar_gecko-inspired_2022}. However, these directional adhesion mechanisms are sensitive to wear and tear and require periodic cleaning for reliable adhesion. On the other hand, insects such as caterpillars and cockroaches are able to climb vertical surfaces with small spines attached to their legs. These microspine grippers increase shear force and traction by engaging with surface asperities. They do not penetrate surfaces, but rather reinforce surface interaction by gripping onto jagged, microscopic edges. Biological studies have looked at the relationship between microspine arrays and surface interaction, comparing the efficacy on different smooth and rough surfaces \cite{dai_roughness-dependent_2002}. The outcome from these studies encourage the design of synthetic microspine arrays to enable maneuverability of uneven terrains, traversability of inclines, and climbing walls \cite{asbeck_scaling_2006, asbeck_designing_2012, iacoponi_simulation_2020, wang_design_2017, wang_palm_2016, jiang_stochastic_2018, sangbae_kim_spinybotii_2005, wang_spinyhand_2019, zi_mechanical_2023, asbeck_climbing_nodate, liu_novel_2019}. Despite the promise, design and control of these arrays can become very complicated, and there exists a trade-off between design complexity and surface adaptability for many of these robots as shown in \Fig \ref{fig:adapt}.

\tikzstyle{background rectangle}=[thin,draw=black]
\begin{tikzpicture}[overlay, show background rectangle]
\node[anchor=south, 
      align=center, yshift=-1.5cm, text width=0.8\textwidth, font =\small]
     {The multimedia attachment can be accessed at https://youtu.be/U9QPgaXc8IU?si=zxycAdzoQ9YBXSA3.}; 
\end{tikzpicture}

Climbing and crawling rigid robots developed in the past decade employ similar `microspine' methods to combat slipping along vertical walls \cite{asbeck_climbing_nodate, li_structure_2020, martone_enhancing_2019, ruotolo_load-sharing_2019, xu_rough_2016}. For example, JPL's LEMUR IIB, a four-limbed robot, features multiple grippers, each equipped with at least 250 microspines capable of adhering to both convex and concave surfaces. Here, each microspine is suspended with a steel hook, allowing it to stretch and move relative to its neighbors to find suitable surfaces to grip. Conceptually, the shear forces from the microspines integrated in the gripper gives the ability to support heavy loads \cite{parness_gravity-independent_2013} and  the necessary traction to even navigate microgravity environments \cite{parness_lemur_2017}. Mechanically, the microspines latch into small asperities on a rock’s surface such as pits, cracks, slopes, or any kind of topography that will snag a hook. Consequently, the success of an individual spine catching is dependent on surface roughness, surface geometry, and asperity distribution \cite{parness_maturing_2017}. 

\begin{figure}[!t]
    \centering
    \includegraphics[width=0.7\linewidth]{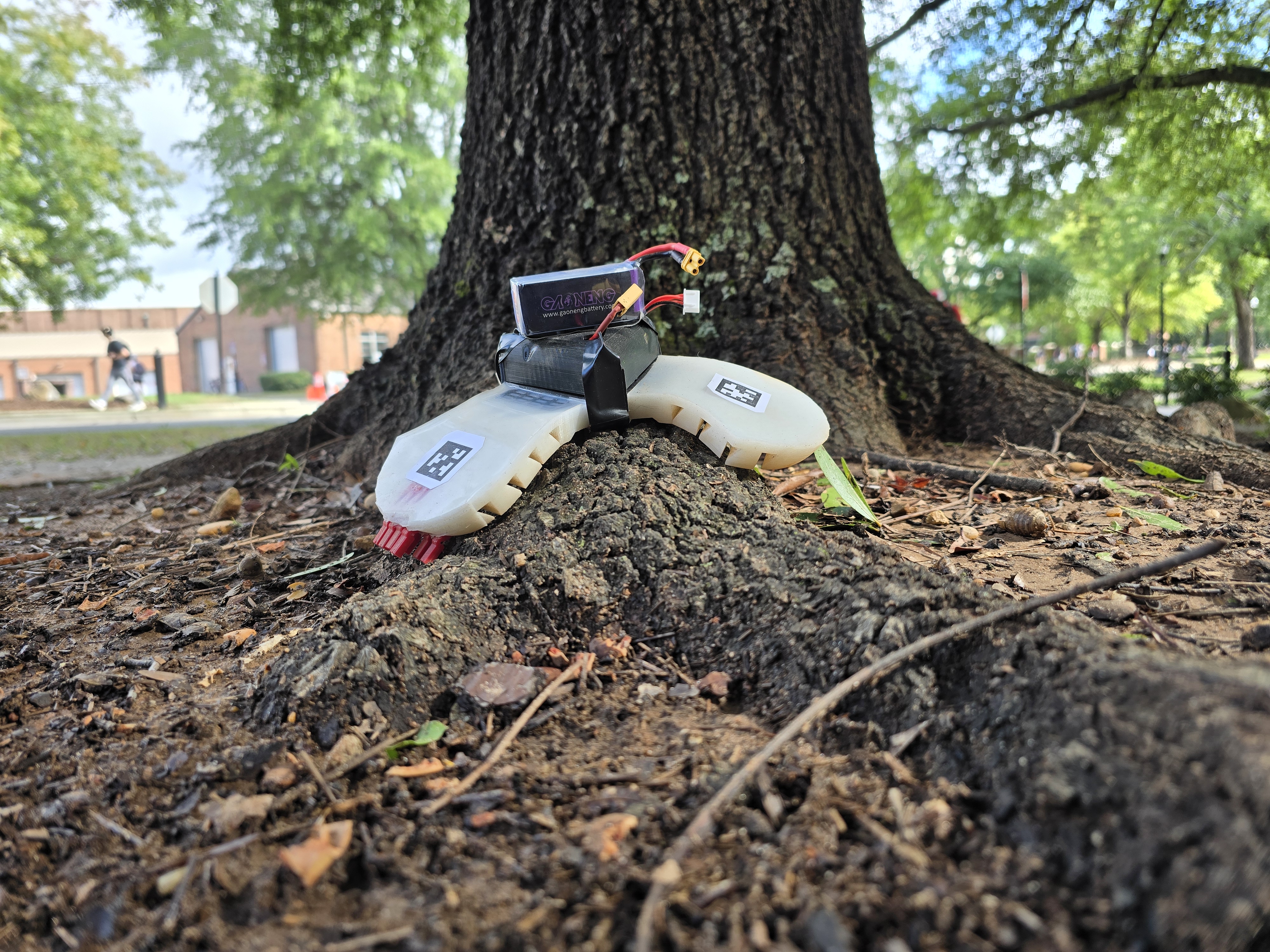}
    \caption{Conformable nature of soft limbs enable traversal over large tree root present on forest floor surface.}
    \label{fig:tree}
\end{figure}

Integrating microspines into Soft Robots (SoRos) is an attractive option due to their adaptability and conformability to changing surface topologies, highlighted in \Fig \ref{fig:tree}. 
The continuum nature and impact resistance of soft materials passively allow SoRos additional flexibility and more effective interaction with complex and non-uniform surfaces. 
However, SoRos lack grip stability, contributing to them historically struggling with efficient locomotion as well as locomoting over unstructured terrain. Because of this, SoRo designs that can traverse outside and perform real tasks outside of a lab setting are under-researched. Integrating microspines has the potential to improve traction, increase grip stability, and provide the ability to reliably maneuver real-world terrains. One of the main design challenges pertains to integrating a soft, low stiffness body with hard, high stiffness micropsines. ``Soft microspines'' made of rubber along the ventral side of the body increase anisotropic friction and adhere using an array of dual material \cite{ta_design_2018}. Similarily, tube feet reminiscent of microspines along the entire underside of the five-limbed star-fish inspired soft robot use similar technique. Locomotion adaptability can be improved by magnetizing the soft tube feet \cite{yang_starfish_2021}. 
Embedding hard objects in soft materials through intelligent mechanical design is critical to take advantage of the benefits of hard spines without hindering the soft deformable properties. A soft inchworm design attaches an array of microspines to either foot of the inchworm using adhesive bonding technology \cite{hu_inchworm-inspired_2019}. However, deeply irregular surfaces remain difficult if not impossible to overcome due to the uniform distribution of the microspines and integration technique that restricts the usage to surfaces with regular, fine asperities. All this calls for need of compliance and independent movement per microspine to increase surface geometry traversability.


\textit{Contributions.} The research addresses the real-world implementation gap by incorporating compliant microspines into the limbs of the Motor Tendon Actuated (MTA) SoRo, significantly enhancing grip stability and traversability across various terrains for mobile SoRos.  The proposed single-material design with intelligent soft-compliant integration reduces complexity by using a single actuator to passively control an entire array fixed to the end of a soft limb. This research, 
(1) proposes a compliant mechanism, two-row stacked microspine array design that improves grip stability and increases traversable surface topologies of mobile SoRos;
(2) identifies critical design parameters that improve locomotion capabilities while reducing complexity by controlling an entire microspine array with only a single actuator through intelligent soft-compliant integration;
(3) investigates the grip stability and repeatability of a baseline SoRo compared against two different microspine array configurations on uniform concrete, partially uniform brick, granular compact sand, and non-uniform tree roots; and
(4) experimentally concludes that the inclusion of compliant microspine arrays in SoRos increases planar displacement on all surfaces through enhanced surface engagement resulting in capabilities to traverse complex, unstructured environments.

\textit{Paper Organization.} The next section explores critical design parameters of the compliant mechanism microspine array and the SoRo prototype. The third section discusses the three experimental prototypes, experimental setup, and pose estimation algorithm. The fourth section highlights the results, validating the microspine array design. The fifth section concludes the paper and discusses the future work.
\section{MICROSPINE ARRAY AND ROBOT DESIGNS}
Mechanically, the critical design parameters of the mechanism that impact the effectiveness of interaction of the microspines with the environmentcan be identified as (1) compliance of individual microspines and the angle of their interaction with a surface, (2) the array configuration involving multiple microspines, and (3) effective integration with the robot body to manage the rigid-soft stiffness mismatch. The other parameters that are out of the hands of the designer, but also highly impact surface engagement, include surface roughness, distribution of asperities, and size of asperities.

\subsection{Compliant Mechanism Microspine Design}
\begin{figure}[h]
    \centering
    \includegraphics[width=\linewidth]{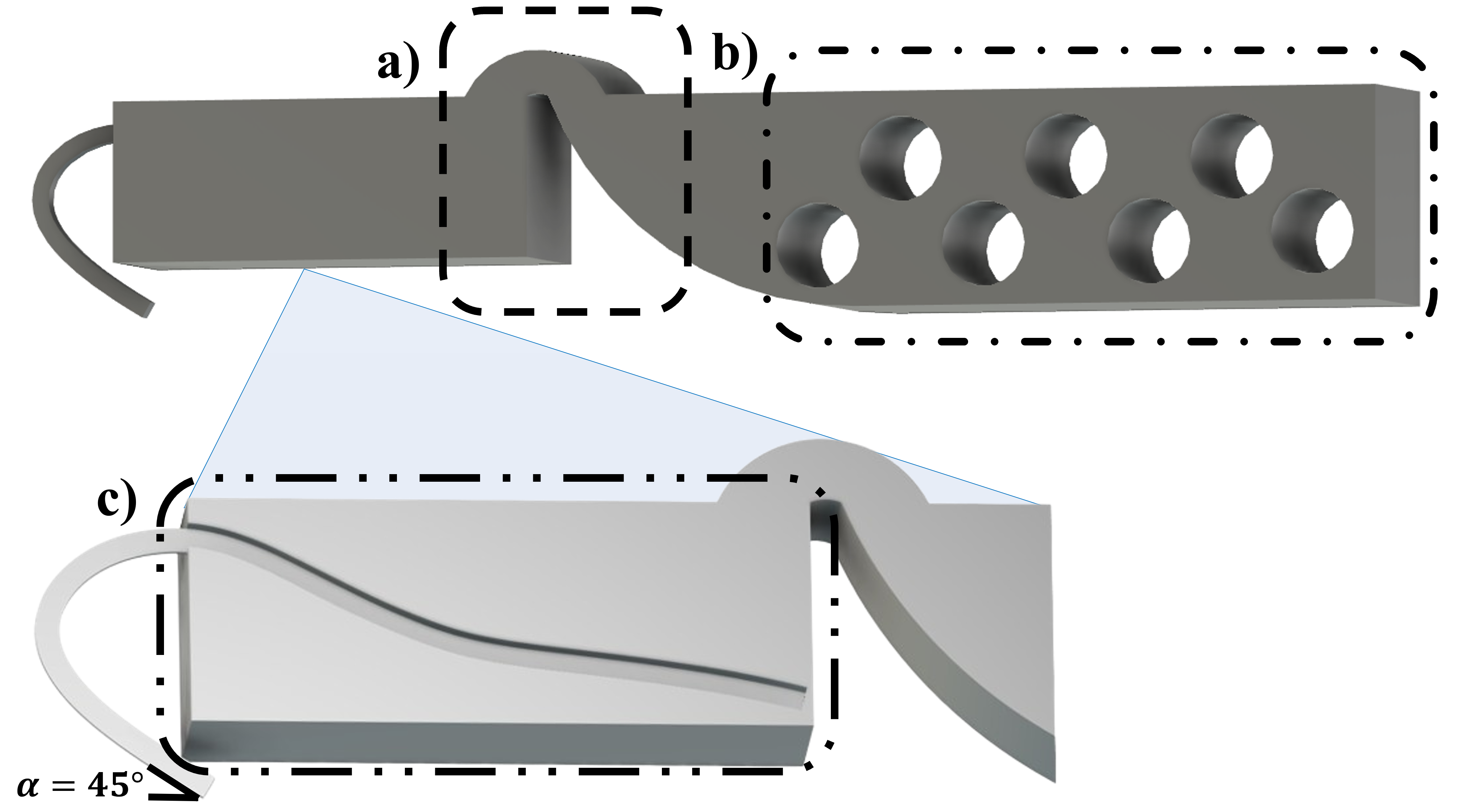}
    \caption{Compliant microspine mechanism design. a) A hinge joint enables passive compliance.  b) Holes embedded on the righthand side facilitate mechanical integration into the soft limb. c) A rigid microspine is inserted in a center channel matching the spine topology set halfway into the mechanism with contact angle of $\alpha=45\degree$}
    \label{fig:comp}
\end{figure}

The single-material mechanism, shown in \Fig \ref{fig:comp}, allows compliance with an exposed joint while simplifying the fabrication process over others presented in the literature. The compliant mechanism is fabricated with an FDM 3D printer and TPU with 95A Shore hardness. Halfway through the additive manufacturing process, the print is paused. The microspine is inserted into a channel left in the middle of the mechanism, highlighted in \Fig \ref{fig:comp}c), and the print is resumed. Once finished, the angle of the bare microspine can be modified for different surface topologies while the body remains secure in the mechanism. The angle of surface interaction, $\alpha$, was fixed at roughly $45^{\circ}$ during testing.

\subsection{Microspine Array Configuration}
\begin{figure}[h]
    \centering
    \includegraphics[width=\linewidth]{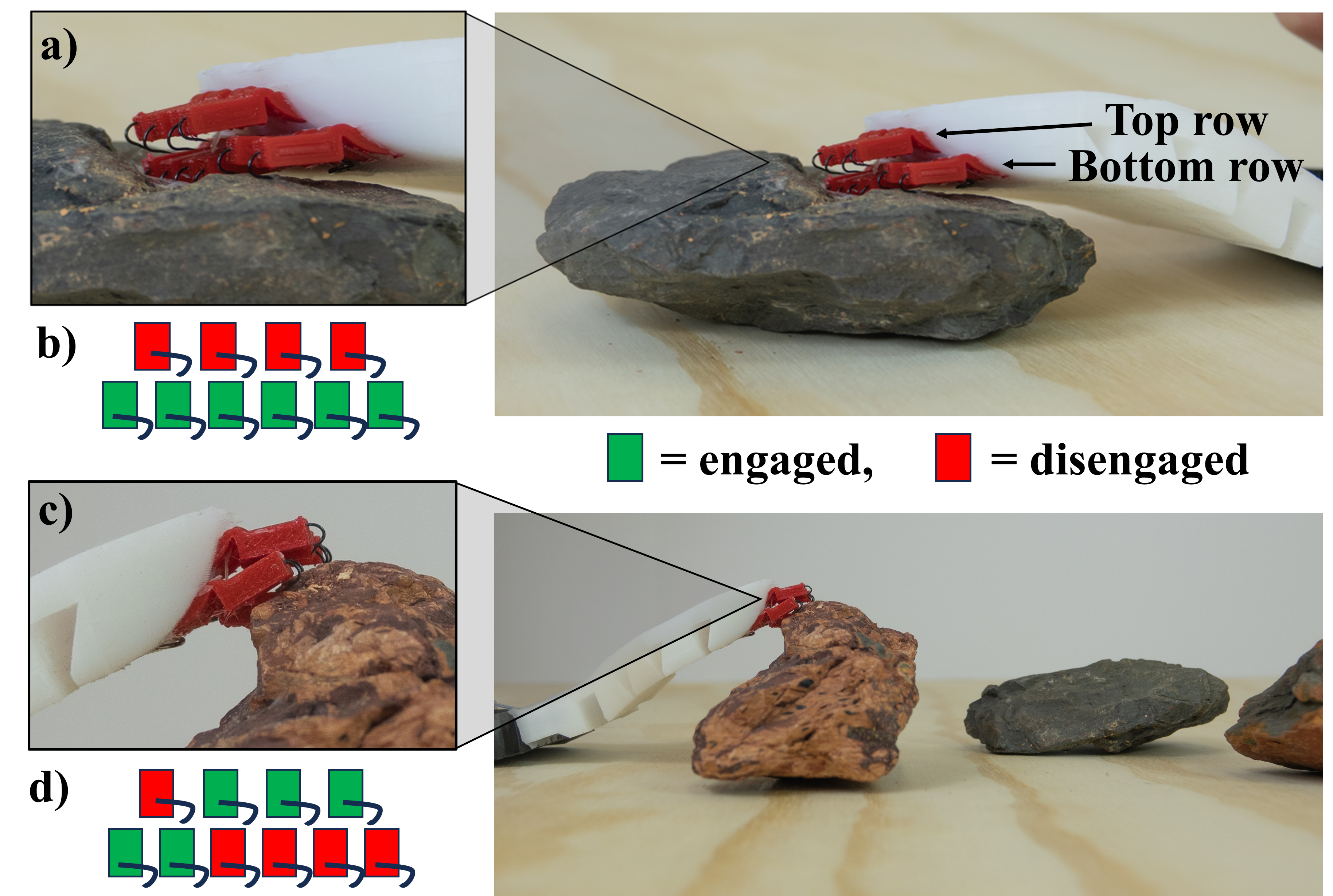}
    \caption{Two-row, stacked array configuration. a) Close-up of the microspines gripping onto a non-uniform rock. b) where all six of the microspines on the bottom row are engaged (green) while the top row remains disengaged (red). c) Close-up of the microspines gripping onto a steeper rock. d) where two of the bottom row and three of the top row microspines are engaged with the surface.}
    \label{fig:eng}
\end{figure}

The array configuration ensures multiple microspines remain active on various complex surfaces. The proposed design comprises of two-row, stacked array configuration consisting of ten microspines with four on the top row and six on the bottom. Experimentally, it has been observed that the microspines on the bottom row are commonly active on more uniform terrain. The top row can become active on steep/highly irregular surfaces without hindering the movement of the bottom row of microspines. The critical parameter when designing the array configuration is ensuring adequate surface interaction and gripping regardless of topology. Crucially, not all microspines need to interact with a surface for the microspine array to be effective, shown in \Fig \ref{fig:eng}. This is a byproduct of the passive nature and built-in redundancy of the system.

\subsection{Effective Soft-Compliant Integration Through Anchoring}
\begin{figure}[h]
    \centering
    \begin{subfigure}{0.4\linewidth}
         \centering
         \includegraphics[width=0.7\textwidth]{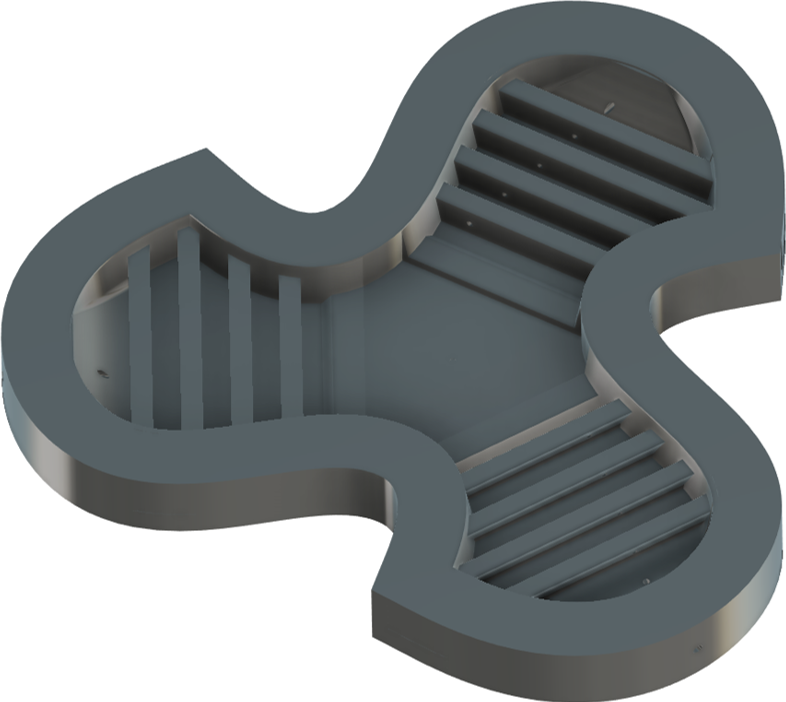}
         \caption{Mold}
         \label{fig:base}\vspace{-20pt}
     \end{subfigure}
     \begin{subfigure}{0.55\linewidth}
         \centering
         \includegraphics[width=0.7\textwidth]{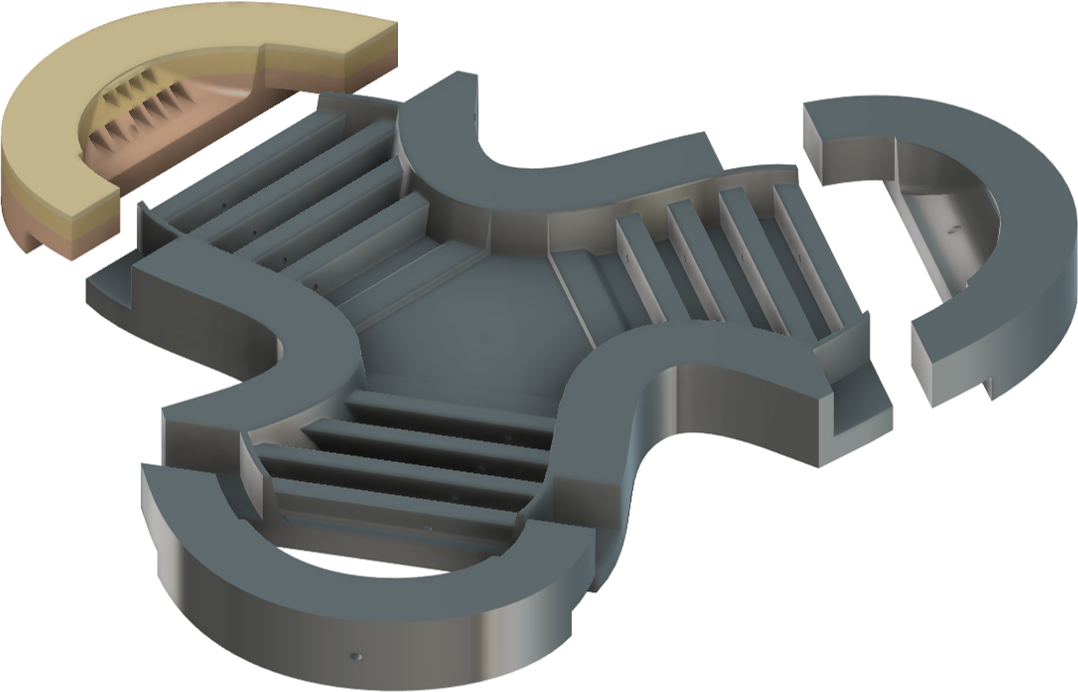}
         \caption{Modified mold}
         \label{fig:one}\vspace{-20pt}
     \end{subfigure}
     \begin{subfigure}{0.45\linewidth}
         \centering
         \includegraphics[width=0.7\textwidth]{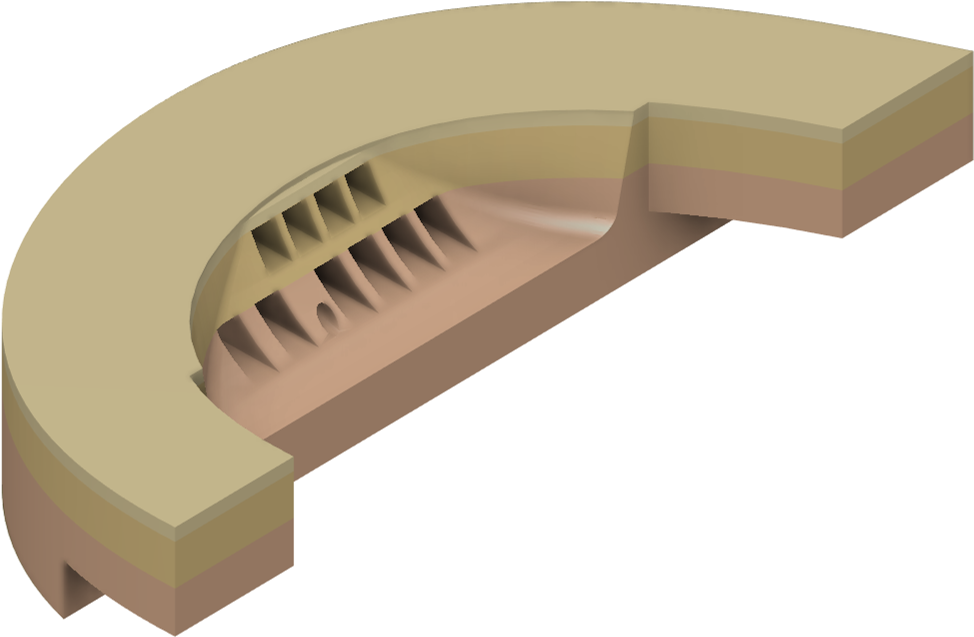}
         \caption{Microspine mold tip}
         \label{fig:two}
     \end{subfigure}
     \begin{subfigure}{0.45\linewidth}
         \centering
         \includegraphics[width=0.7\textwidth]{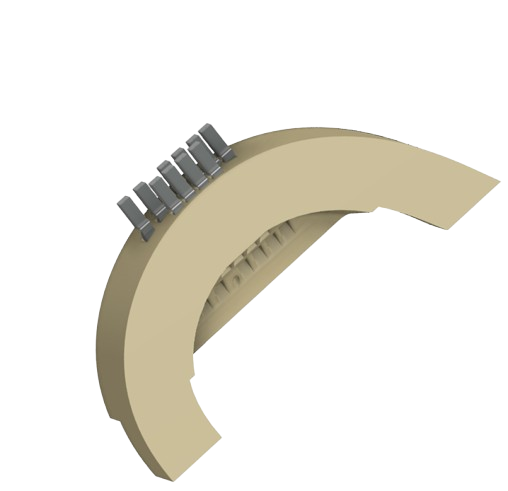}
         \caption{Integrated microspines}
         \label{fig:two}
     \end{subfigure}
    \caption{Modular ends of the mold enable soft-rigid integration. a) A baseline robot mold. b) A modified mold that allows different configurations of microspine arrays per limb. c) Microspine compatible end mold with holes for a two-row stacked microspine array configuration. d) End mold with integrated microspine mechanisms and ready for casting.}
    \label{fig:totalMold}
\end{figure}

The soft-compliant integration reduces design complexity by allowing each microspine to passively move independent of one another with a single actuator controlling the entire array configuration, equivalently the limb. To achieve this, a mold is created with channels for each microspine compliant mechanism to attach to the tip of a SoRo limb. The SoRo prototype used for experimentation is cast out of DragonSkin\texttrademark~ silicone rubber using a custom mold, shown in \Fig \ref{fig:totalMold}. Therefore, a modified limb mold is used for integrating the microspine array in a consistent, standardized manner. Half of the compliant mechanism contains holes that mechanically anchor it into the silicone limb, highlighted in \Fig \ref{fig:comp}b), preventing it from being freely pulled out of the limb during microspine gripping. This anchoring method is essential for ensuring the microspine does not come loose over time. The remaining, exposed half of the mechanism contains the microspine.

\subsection{Soft Robot Design}
\begin{figure}[!h]
    \centering
    \includegraphics[width=\linewidth]{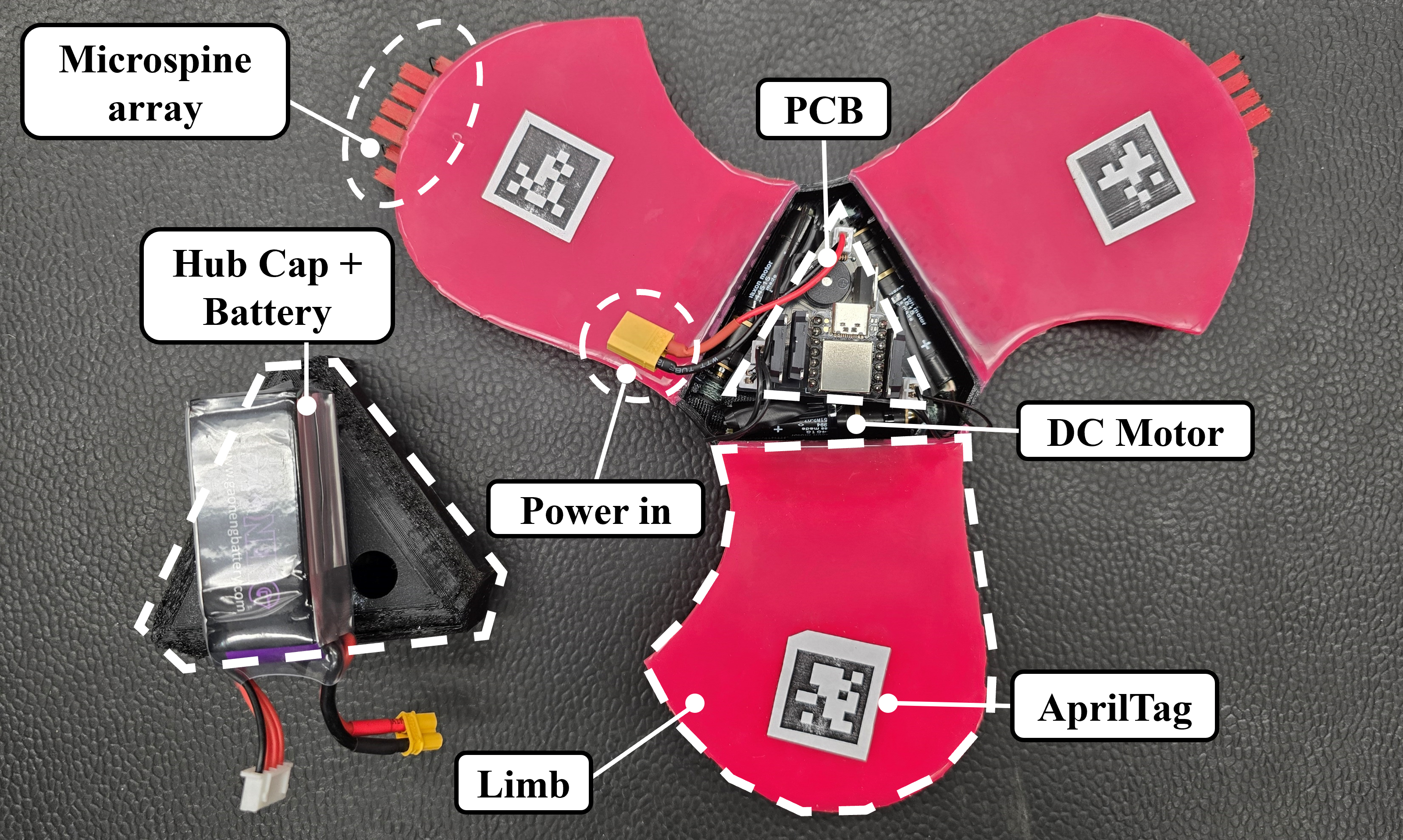}
    \caption{The externally powered three-limb SoRo contains soft material limbs and a flexible, central hub that houses DC motors and a custom-designed PCB. The three AprilTags on the limbs help with pose tracking during experiments.}
    \label{fig:mech}
\end{figure}

A tetherless, three limb MTA SoRo with on-board power and processing with AprilTags on each limb is used as the experimental prototype. The components of the physical robot are shown in \Fig \ref{fig:mech}. 
Outward trapezoid cavities are introduced on the underside of each limb to provide optimal stiffness and curling ability. This allows the robot to lift the limb and electronic payload. The use of MTA for body deformation enables reliable and efficient limb actuation. The reader may refer to \cite{freeman_topology_2023} for more details about the topology and morphology design of such modular reconfigurable soft robots.

\section{EXPERIMENTATION}

\subsection{Experimental Configurations}
\begin{figure}[h]
    \centering
    \begin{subfigure}{0.32\linewidth}
         \centering
         \includegraphics[width=0.8\textwidth]{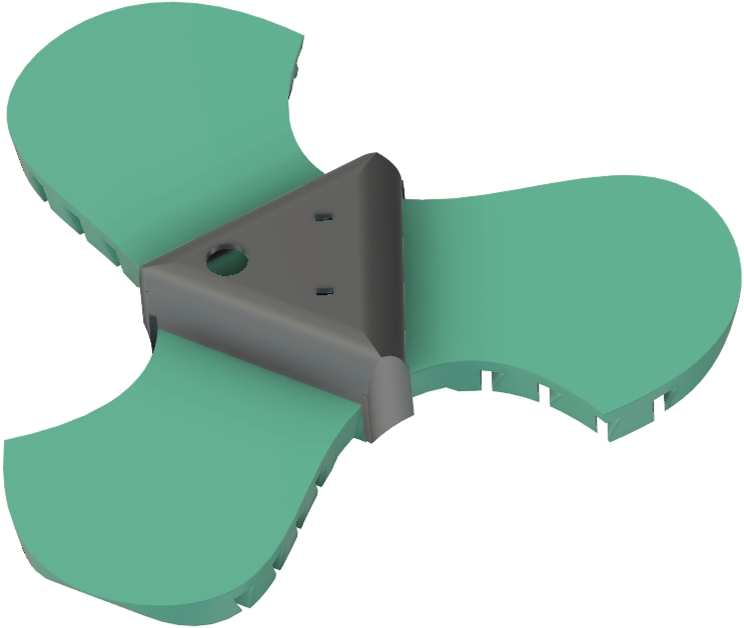}
         \caption{0ML}
         \label{fig:base}
     \end{subfigure}
     \begin{subfigure}{0.32\linewidth}
         \centering
         \includegraphics[width=0.8\textwidth]{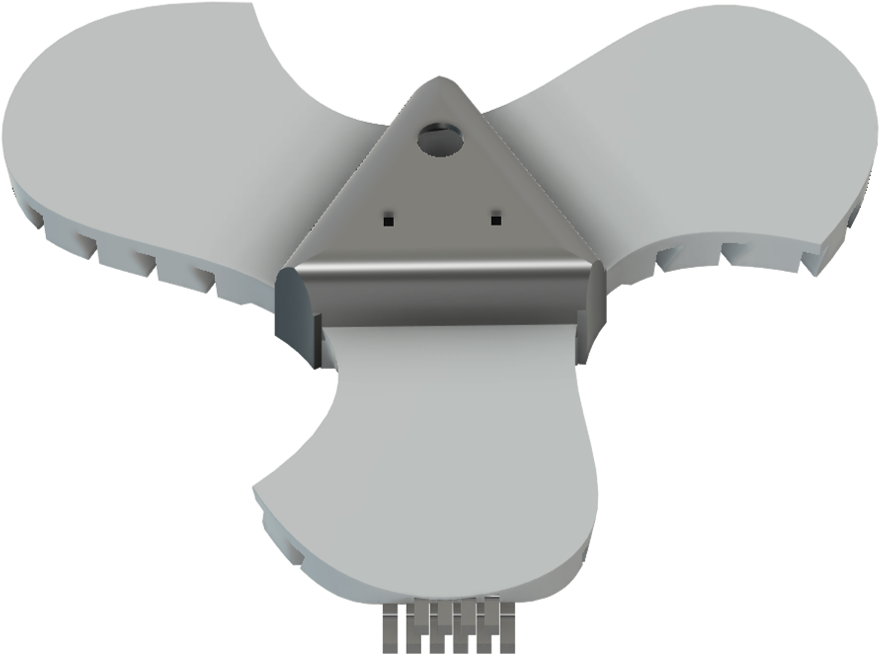}
         \caption{1ML}
         \label{fig:one}
     \end{subfigure}
     \begin{subfigure}{0.32\linewidth}
         \centering
         \includegraphics[width=0.8\textwidth]{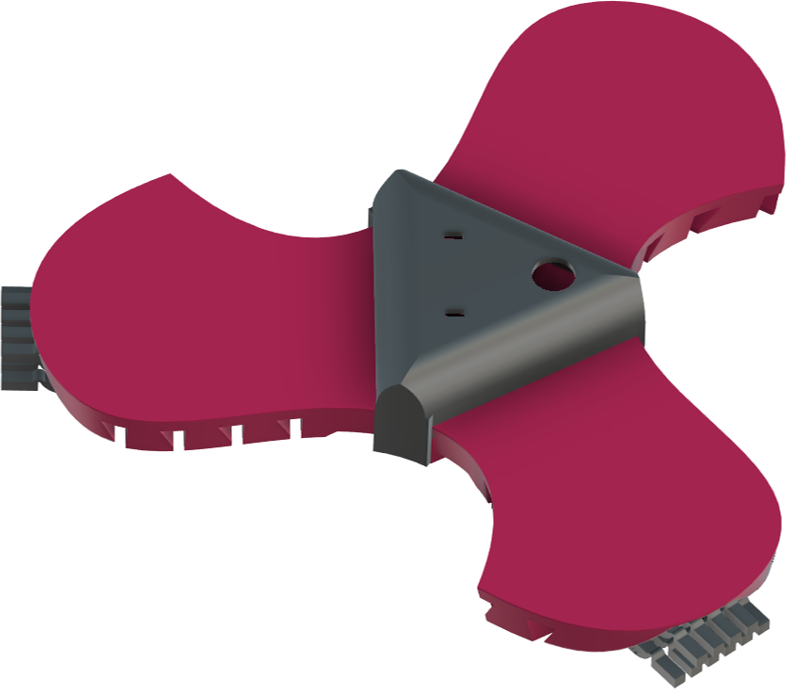}
         \caption{2ML}
         \label{fig:two}
     \end{subfigure}
    \caption{The three different microspine configurations explored in the experiments: a) a baseline aqua SoRo; b) a white SoRo with a total of 10 microspines configured in an array on one limb; c) a red SoRo with a total of 20 microspines configured in two arrays on two limbs.}
    \label{fig:SoRos}
\end{figure}

The baseline experimental configuration with Zero Microspine Limbs, 0ML, is a three-limb MTA SoRo used in  \cite{freeman2024environmentcentriclearningapproachgait}. This is compared against two other configurations. The first is One Microspine Limb, 1ML, with the microspine array affixed to a single limb. The second is Two Microspine Limb, 2ML, with microspine arrays equipped on two limbs. In all these configurations, the microspines are angled towards the surface opposite the expected direction of movement. The CAD models of the three prototypes are shown in \Fig \ref{fig:SoRos}.

\begin{figure}[h]
    \centering
    \includegraphics[width=0.55\columnwidth]{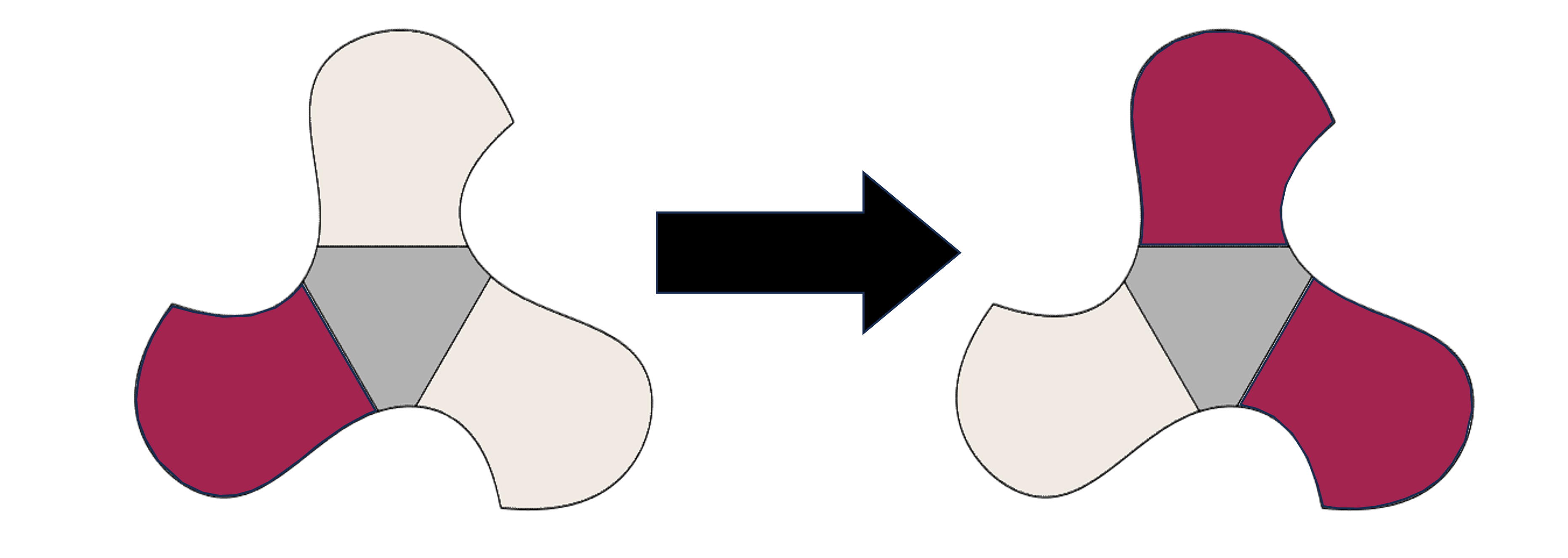}
    \caption{Translation gait: one limb (in maroon) actuates while the other two relax, then the previously relaxed limbs actuate/curl while the first relaxes.}
    \label{fig:trans}
\end{figure}

A push-pull translation gait, \Fig~\ref{fig:trans}, is used for the experiments where the gait sequence involves actuation of one limb followed by actuation of the two relaxed limbs. For this locomotion gait, both 0ML and 1ML maintain the same starting pose for a desired path as the single limb that is first actuated in each robot aligns with the intended direction of movement. However, the starting orientation for 2ML is flipped by $180\degree$ so that the two limbs that actuate together face forward. In this configuration, the center point equidistant from the tips of the two microspine limbs faces the intended direction of movement instead of a single limb. 
This gait was used on all surfaces for all the three robot configurations to facilitate fair comparison. However, an improved performance is possible with an optimized gait that can be synthesized for each robot configuration and surface combination. This may be explored through gait discovery described in \cite{freeman2024environmentcentriclearningapproachgait}.

\subsection{Experimental Setup and Field Experiments}
\begin{figure}[h]
    \centering
    \includegraphics[width=\linewidth]{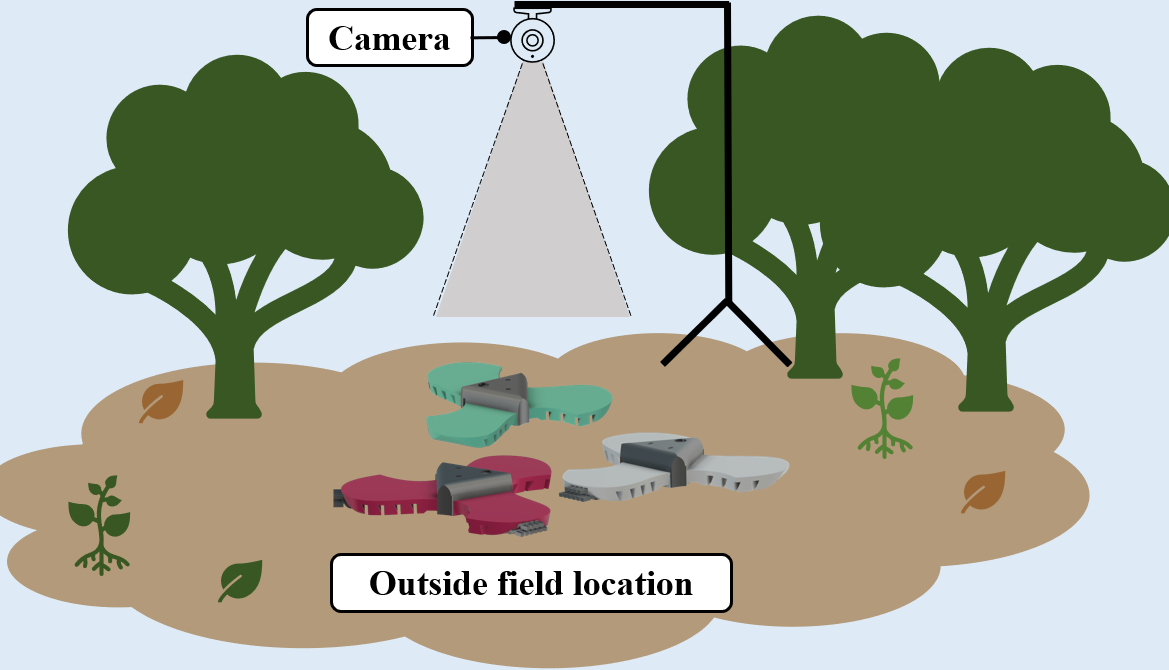}
    \caption{Example of experimental setup for outside field locations in a forest with a dirt patch. The overhead camera is orthogonal to the ground and approximately 4' above the surface.}
    \label{fig:exp}
\end{figure}

The experiments are carried out with an overhead camera attached to a tripod next to the test area, shown in \Fig \ref{fig:exp}.  The starting pose, position and orientation, is same for each SoRo to ensure consistency. Tracking is performed with an AprilTag fixed to the end of each limb serving as fiducial markers. The average displacement per gait as well as overall displacement for each trial run is recorded.

The field experiments are performed on four surfaces of increasing roughness and unstructured nature - uniform concrete, partially uniform brick, granular compact sand with pebbles, and a non-uniform forest floor containing leaf litter and large tree roots. 
Three trials (60 gaits per trial) are performed for each configuration and surface, resulting in a total of 36 trials. A 36h11 family AprilTag from the AprilTag visual fiducial system \cite{Olson_2011} is attached to each limb to aid in tracking. The tracking algorithm and data processing code is available at
\href{https://github.com/AgileRoboticsLab/SoftRobotics-Microspines}{github.com/AgileRoboticsLab/SoftRobotics-Microspines}.

\section{RESULTS}
\begin{figure}[!h]
    \centering
    \includegraphics[width=\linewidth]{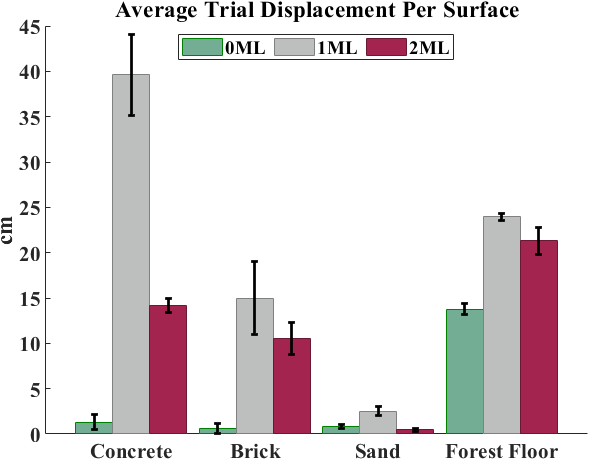}
    \caption{The average displacement data per prototype and surface combination. The standard deviation is indicated by error bars.}
    \label{fig:data}
\end{figure}

The translation results of 0ML, 1ML, and 2ML are compared on the four experiment surfaces. 
\Fig \ref{fig:data} shows the average displacement and standard deviation of the three trials (60 gaits/trial = 180 gaits) for each robot configuration. 

The concrete surface is level with a uniform distribution of asperities. Here, both 1ML and 2ML outperform 0ML in terms of displacement. In fact, 1ML had an average displacement of 39.63cm which $>30\times$ greater than travelled by 0ML, 1.32cm. 2ML traveled 14.13cm, $>10\times$ as much as 0ML. 2ML had higher consistency across trials, having a standard deviation of 0.76 whereas 0ML's standard deviation was 0.83.

The next surface of partially uniform brick contains gaps in between bricks that can cause microspines to become stuck. This was observed to happen randomly across all trials. However, in every instance where this occurred, given the soft and deformable nature of the soft robot, the stuck microspine were able to wiggle free, enabling continued movement of the robots. Despite these difficulties, both 1ML (traveling 15cm) and 2ML (traveling 10.52cm) had far greater average displacement than 0ML (0.59cm), roughly $25 \times$ and $18 \times$ more, respectively. This was the only surface where 0ML had the lowest `gait' standard deviation, which can be attributed to the other two robots randomly getting stuck when crossing over brick perimeters.  

The third surface of granular, compact sand was not entirely level and various pebbles, holes, insects, and small sticks were scattered around the testing area. This surface was difficult to overcome for all three robots as the compact sand was covered in a loose, granular top layer that was easy to become partially submerged in. Due to this, the average displacement is far lower on this surface. The two microspine limbs on 2ML seemed to dig itself deeper, resulting in average displacement of only 0.46cm and the lowest standard deviation. 1ML still outperformed 0ML with over 3 times increased displacement, traveling 2.53cm compared to 0.82cm.

\begin{figure}[!h]
    \centering
    \includegraphics[width=\linewidth]{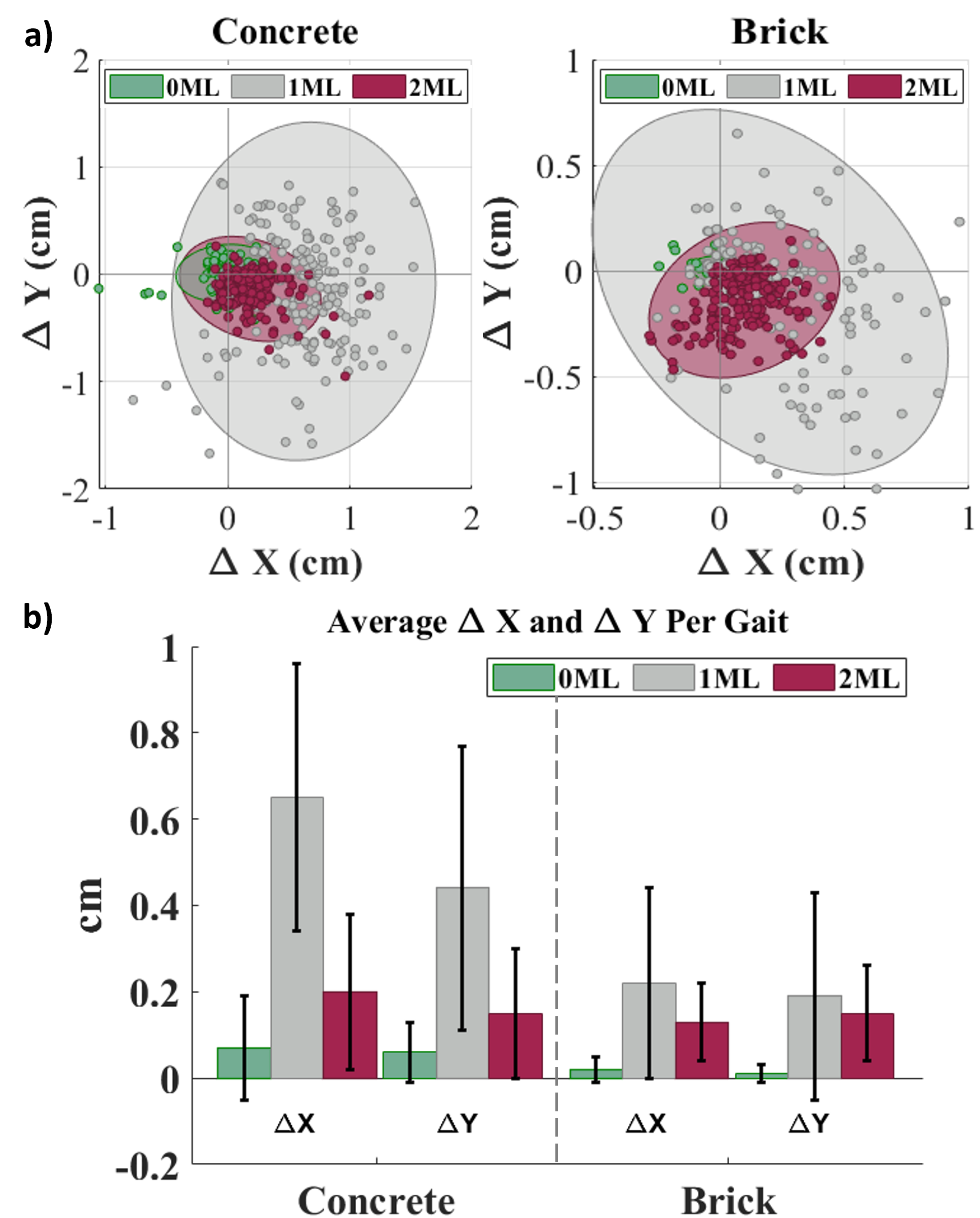}
    \caption{Gait analysis. a) Every $\Delta$X and $\Delta$Y position per gait on concrete and brick. b) The average absolute $\Delta$X and $\Delta$Y displacement per gait on concrete and brick.}
    \label{fig:delta}
\end{figure}

The final surface of forest floor surface was completely unstructured and non-uniform with highly varying terrain. Specifically, the prototypes had to first overcome a large, 4" tall tree root and then move through leaves, acorns, and other tree debris. All SoRo configurations were able to successfully traverse over the cylinder-like tree root due to the conformable nature of the soft limbs, highlighted in \Fig \ref{fig:tree}. However, only the 1ML and 2ML were able to navigate through the thick tree debris after making it over the large tree root; 0ML became stuck at the base of the tree root in each of its three trials. This is exhibited in the average displacement of 13.74cm for 0ML, 23.95cm for 1ML, and 21.32cm for 2ML. The forest floor was critical for testing as it was the most unstructured of the four experimental surfaces and showcases the benefits of the soft limbs paired with the added gripping stability of the microspine array.

For each of the 36 trials, the robots move for 60 gaits, resulting in a total of 2,160 gaits or 720 gaits per robot. Each gait results in relative change in position in the robot coordinate system $\Delta X, \Delta Y$. The locomotion consistency can be analyzed by examining the 180 poses on a given surface per prototype, such as those shown in \Fig \ref{fig:delta}a). The average displacement per gait ($\Delta X,\Delta Y)$  as well as the standard deviation over all gaits per prototype/surface combination is visualized in \Fig \ref{fig:delta}b). Only the uniform/partially uniform surfaces were analyzed as the gait-to-gait movement is much less consistent on the other two surfaces due to non-uniformity. On concrete, the average gait displacement per gait of both 1ML (0.65cm,0.44cm) and 2ML (0.20cm,0.15cm) is greater than 0ML (0.07cm,0.06cm). On brick, both 1ML (0.22cm,0.19cm) and 2ML (0.13cm,0.15cm) outperform 0ML (0.02cm,0.01cm). The relative standard deviation (RSD), is defined by the following equation:

\begin{equation*}
    RSD = \frac{\text{standard deviation}}{\text{mean}} \times 100
\end{equation*}

The RSD on concrete is 135.96\% for 0ML, 37.05\% for 1ML, and 70.55\% for 2ML. On the brick surface, the RSD is 159.94\% for 0ML, 54.24\% for 1ML, and 89.95\% for 2ML. Both 1ML and 2ML have a significantly lower RSD than 0ML, indicating the microspine array provides greater grip stability through improved repeatability.

\begin{figure*}[htbp]
    \centering
    \includegraphics[width=0.95\textwidth]{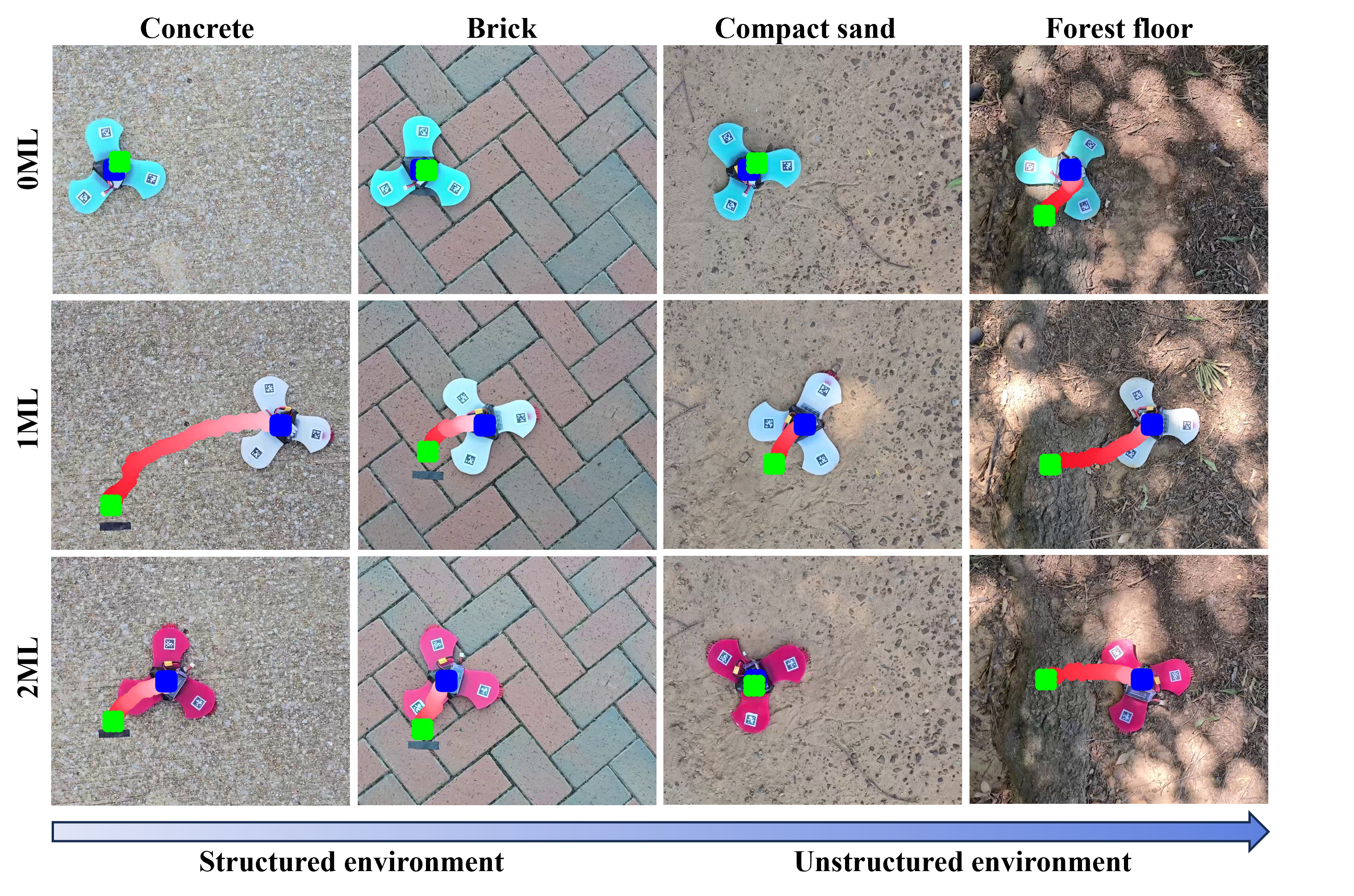}
    \caption{Experimental results for 0ML, 1ML, and 2ML on concrete, brick, compact sand, and a forest floor. Rows represent the different surfaces increasing in unstructured nature with the three different prototypes distinguished by columns.}
    \label{fig:surfaces}
\end{figure*}

Examples of a single trial of each prototype row on each surface column is visualized in \Fig \ref{fig:surfaces} with additional data in the accompanying video. The starting and end points are green and blue dots, and the path of traversal is a red, gradient line. On all surfaces, 1ML interacts with the environment significantly more than the baseline 0ML, resulting in greater overall movement. On every surface except for compact sand, 2ML outperforms 0ML. 

\section{CONCLUSION AND FUTURE WORK}
SoRos have shown immense potential with inherent conformability and adaptability to a multitude of surfaces, yet they previously lacked adequate grip stability to overcome non-uniform surfaces present outside of lab environments. Improving environment interaction using compliant microspines is one of the missing piece that will facilitate shrinking of the real-life realization gap. The proposed compliant microspine design resolves the stiffness mismatch through soft-compliant integration technique. The stacked array configuration enables the SoRo to maintain surface interaction when extreme surface discrepancies are present. The results from field experiments reflect the improved performance (grip stability and repeatability) of two microspine array configurations over a baseline SoRo on four different, ruggedized surfaces. 
They confirm that microspines are a vital technology for increasing terrain traversability in mobile SoRos. Future work includes optimizing microspine array configurations for different surfaces, performing additional field experiments, and exploring the generalizability of the design to different prototypes.

\section*{ACKNOWLEDGMENT}

We thank Bek Ervin for fabricating the 0ML prototype.



\bibliographystyle{IEEEtran}
\bibliography{IEEEabrv,MicroSpine1_NoLink.bib}
\end{document}